\def\year{2019}\relax
\documentclass[letterpaper]{article} 
\usepackage{aaai19}  
\usepackage{times}  
\usepackage{helvet}  
\usepackage{courier}  
\usepackage{url}  
\usepackage{graphicx}  
\usepackage{amsmath}
\usepackage{algorithm}
\usepackage{booktabs}
\usepackage{subfigure}

\usepackage{algorithm}
\usepackage{mathrsfs}
\usepackage{amsfonts,amssymb,amsthm,subfigure}
\usepackage{epsfig}
\usepackage{graphicx}
\usepackage{setspace}
\usepackage{color}
\usepackage{threeparttable}
\usepackage{Dcolumn}
\usepackage{multirow}
\usepackage{ upgreek }
\usepackage[table]{xcolor}
\usepackage{subfigure}
\usepackage{setspace}
\usepackage{amsmath}
\usepackage{algpseudocode}
\usepackage{CJK}
\usepackage{algorithmicx}
\begin{document}
%
\title{DT-LET: Deep Transfer Learning by Exploring where to Transfer}
\author{Jianzhe Lin, Qi Wang, Rabab Ward, Z. Jane Wang
}
\maketitle
\begin{abstract}
Previous transfer learning methods based on deep network assume the knowledge should be transferred between the same hidden layers of the source domain and the target domains. This assumption doesn't always hold true, especially when the data from the two domains are heterogeneous with different resolutions. In such case, the most suitable numbers of layers for the source domain data and the target domain data would differ. As a result, the high level knowledge from the source domain would be transferred to the wrong layer of target domain. Based on this observation, ``where to transfer" proposed in this paper should be a novel research frontier. We propose a new mathematic model named DT-LET to solve this heterogeneous transfer learning problem. In order to select the best matching of layers to transfer knowledge, we define specific loss function to estimate the corresponding relationship between high-level features of data in the source domain and the target domain. To verify this proposed cross-layer model, experiments for two cross-domain recognition/classification tasks are conducted, and the achieved superior results demonstrate the necessity of layer correspondence searching.

\end{abstract}

\section{Introduction}
Transfer learning or domain adaption aims at digging potential information in auxiliary source domain to assist the learning task in target domain, where bare labeled data with prior knowledge exist \cite{Survey}. Without the help of related source domain data, the learning tasks like image classification or recognition would fail with insufficient pre-existing labeled data. For most big data problems, the labeled data are highly required but always not enough as labeling process would be quite tedious and laborious. Therefore, having a better use of auxiliary source domain data by transfer learning methods has attracted researchers' attention.

It should be noted direct application of labeled source domain data to a new scene of target domain would result in poor performance due to the semantic gap between the two domains, even they are representing the same objects \cite{Heterogeneous2011zhu}\cite{Learning2012Lixin}. The semantic gap can be resulted from different acquisition conditions(illumination or view angle) and the use of different cameras or sensors. Transfer learning methods are proposed to overcome this distribution divergence or feature bias \cite{Translated2009Dai}\cite{Liu2017IJCAI}\cite{Balanced2017Jingdong}. Traditionally, these transfer learning methods would adopt linear or non-linear transformation with kernel function to learn a common subspace on which the gap is bridged \cite{Learning2017Yugang}. Recent advancement has proven that the features learnt on such common subspace are inefficient. Therefore, deep learning based model has been introduced due to its power on high level feature representation.

Current deep learning based transfer learning topics include two research branches, what knowledge to transfer and how to transfer knowledge \cite{Gopalan2015visual}. For what knowledge to transfer, researchers mainly concentrate on instance-based transfer learning and parameter transfer approaches. Instance-based transfer learning methods assume that only certain parts of the source data can be reused for learning in the target domain by re-weighting \cite{Liu2016ICML}. As for parameter transfer approaches, people mainly try to find the pivot parameters in deep network to transfer to accelerate the transfer process. For how to transfer knowledge, different deep networks are introduced to complete the transfer learning process. However, for both research areas, the right correspondence of layers is ignored.

For what knowledge to transfer problem, the transferred content might even be negative or wrong. A fundamental problem for current transfer learning work should be negative transfer\cite{Distant2017Tan}. If the knowledge from the source domain to target domain is transferred to wrong layers, the transferred knowledge is quite error-prone. With the wrong prior information added, bad effect can be generated on target domain data. For how to transfer knowledge problem, as the two deep networks for the source domain data and the target domain data need to have the same number of layers, the two models could not be optimal at the same time. This situation is especially important for cross-resolution heterogeneous transfer. For data with different resolutions, The data with higher resolution might need more max-pooling layers than the data with lower resolution, and more neural network layers are needed. Based on the above observation and assumption, we propose a novel research topic, where to transfer. In this work, the number of layers for two domains does not need to be the same, and optimal matching of layers will be found by the newly proposed objective function. With the best parameters from the source domain data transferred to the right layer of the target domain, the performance of the target domain learning task can be improved.

\begin{figure}[!ht]
\centering
\includegraphics[width=\linewidth]{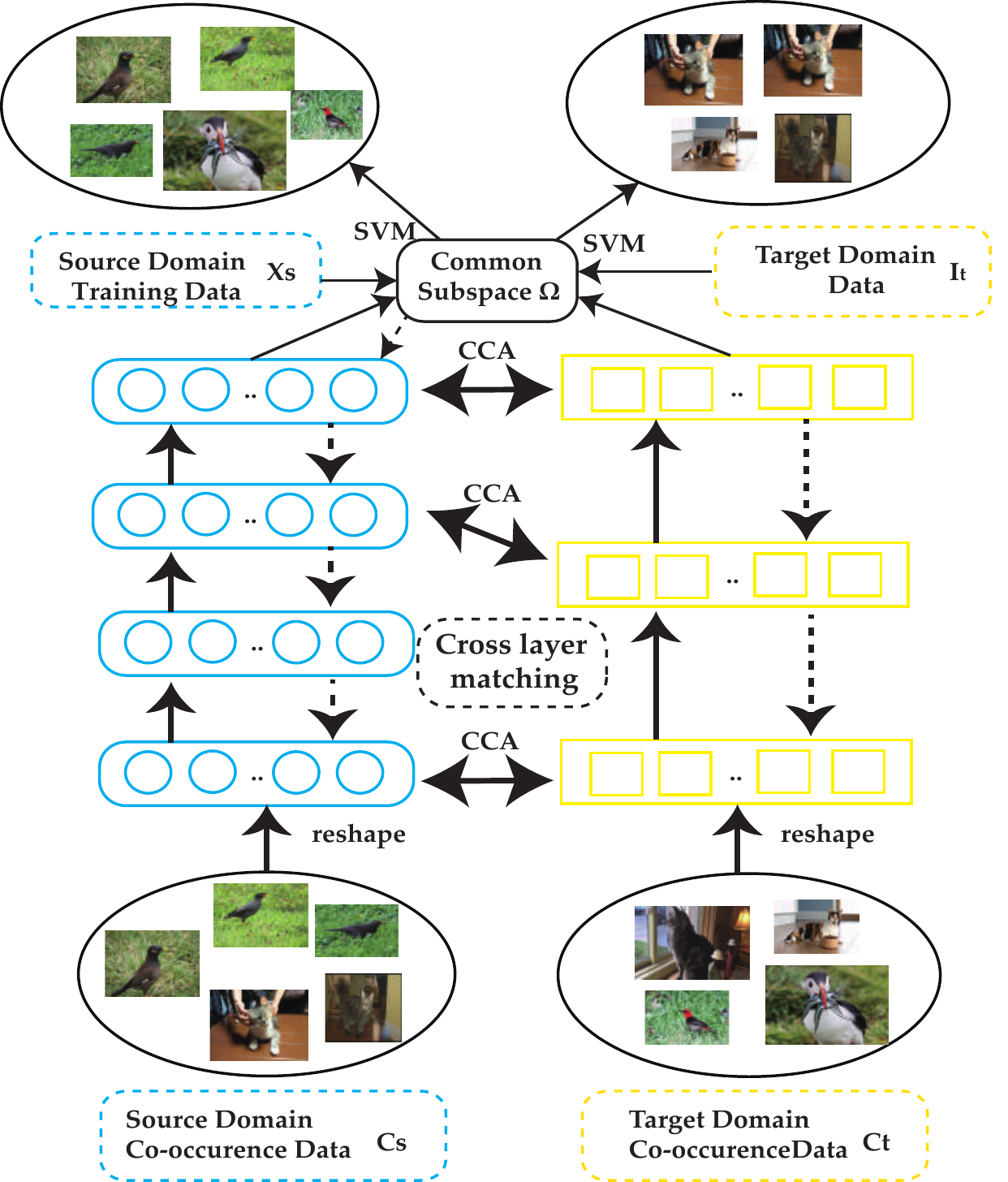}
\caption{The flowchart of the proposed DT-LET framework. The two neural networks are first trained by the co-occurrence data $C_s$ and $C_t$. It is found the best layer matching to match the corresponding layers by CCA. After network training, the common subspace is found and the training data $D_S^l$ is transferred to such space to train SVM classifier, to classify $D_T$.}
\label{fig:1}
\end{figure}

The proposed work is named Deep Transfer Learning by Exploring where to Transfer(DT-LET), which is based on Stacked Auto-Encoders\cite{Auto2018Zhuang}. A detailed flowchart is shown in Fig. \ref{fig:1}. The main contributions are concluded as follows.

\begin{itemize}
  \item This paper for the first time introduces the where to transfer problem. The deep networks from the source domain and the target domain no longer need to be with the same parameter settings, and the cross-layer transfer learning is proposed in this paper.

  \item We propose a new principle for finding the correspondence between neural networks in the source domain and in the target domain by defining new unified objctive loss function. By optimizing this objective function, the best setting of two deep networks as well as the correspondence relationship can be figured out.

\end{itemize}

\section{Related Work}
Deep learning intends to learn nonlinear representation of raw data to reveal the hidden features \cite{Deep2016long}. However, a large number of labeled data are required to avoid over-fitting during the feature learning process. To achieve this goal, transfer learning has been introduced to augment the data with prior knowledge. By aligning data from different domains to high-level correlation space, the data information on different domains can be shared. To find this correlation space, many deep transfer learning frameworks have been proposed in recent years. The main motivation is to bridge the semantic gap between the two deep neural networks of the source domain and the target domain. However, due to the complexity of transfer learning, some transfer mechanisms still lack satisfying interpreting. Based on this consideration, quite a few interesting ideas have been generated. To solve how to determine which domain to be source or target problem, Fabio et al. \cite{Autodial} propose to automatically align domains for the source and target domain. To boost the transfer efficiency and find extra profit during the transfer process, deep mutual learning \cite{deepmutual} has been proposed to transfer knowledge bidirectionally. The function of each layer in transfer learning is explored in \cite{CactusNets}. The transfer learning with unequal classes and data are experimented in \cite{optimal} and \cite{investigating} respectively. However, all the above works still just explain what knowledge to transfer and how to transfer knowledge problems. They still ignore interpreting the matching mechanisms between layers of deep networks of the source domain and the target domain. For this problem, We also name it as DT-LET: Deep Transfer Learning by Exploring Where to Transfer. For this work, we adopt stacked denoising autoencoder(SDA) as the baseline deep network for transfer learning.

Glorot et al. for the first time employ stacked denoising to learn homogeneous features based on joint space for sentiment classification \cite{glorot2010}. The computation complexity is further reduced by Chen et al. by the proposing of Marginalized Stacked Denoising Autoencoder(mSDA) \cite{Chen2012Marginalized}. In this work, some characteristics of word vector are set to zero in the equations of expectation to optimize the representation. Still by matching the marginal as well as conditional distribution, Zhang et al. and Zhuang et al. also develop SDA based homogeneous transfer learning framework \cite{Supervised2015zhuang}\cite{Deep2015zhang}. For heterogeneous case, Zhou et al. \cite{Hybrid2014zhou} propose an extension of mSDA to bridge the semantic gap by finding the cross-domain corresponding instances in advance. Google brain team in recent time introduces generative adversarial network to SDA and propose the Wasserstein Auto-Encoders \cite{Wasserstein} to generate samples of better quality on target domain. It can be found SDA is with quite high potential, and our work also chooses SDA as the basic neural network for the where to transfer problem.

\section{Deep Mapping Mechanism}
The general framework of such deep mapping mechanism can be summarized as three steps, network setting up, correlation maximization, and layer matching. We would like to first introduce the deep mapping mechanism by defining the variables.

The samples in the source domain are denoted as $D_S = \{I_i^s\}_{i=1}^{n_s}$, in which the labeled data in the source domain is further denoted as $D_S^l = \{X_i^s, Y_i^s\}_{i=1}^{n_l}$, they are used to supervise the classification process. In the target domain, the samples are demoted as $D_T = \{I_i^t\}_{i=1}^{n_t}$. The co-occurrence data \cite{Learning2016Yang}(the data in the source domain and the target domain belonging to the same classes but with no prior label information) in the source domain are denoted as $C_S = \{C_i^s\}_{i=1}^{n_c}$, in target domain are denoted as $C_T = \{C_i^t\}_{i=1}^{n_c}$. They are further jointly represented by $D^C = \{C_i^S, C_i^T\}_i^{n_c}$, which are used to supervise the transfer learning process. The parameters of deep network in the source domain are denoted by $\Theta^S = \{W^s, b^s\}$, and $\Theta^T = \{W^t, b^t\}$ in the target domain.

The matching of layers is denoted by ${R_{s,t}} = \{ r_{{i_1},{j_1}}^1,r_{{i_2},{j_2}}^2,...,r_{{i_a},{j_b}}^m\}$, in which a represents the total number of layers for the source domain data, and b represents the total layers for the target domain data. $m$ is the total number of matching layers. We define here, if ${\rm{m}} = \min \{ a - 1,b - 1\} $(as the first layer is the original layer which will not be used to transfer, m is compared with a-1 or b-1 instead of a or b), we define the transfer process as full rank transfer learning; else if ${\rm{m}} < \min \{ a - 1,b - 1\} $, we define this case as non-full rank transfer learning.

The common subspace is represented by $\Omega$ and the final classifier is represented by $\Psi$. The labeled data $D_S^l$ from the source domain are used to predict the label of $D_T$ by applying $\Psi(\Omega(D_T))$.

\subsection{Network setting up}

The stacked auto-encoder (SAE) is first employed in the source domain and the target domain to get the hidden feature representation $H^S$ and $H^T$ of original data as shown in eq. (\ref{eq:1}) and eq. (\ref{eq:2}).

\begin{equation}
\begin{aligned}
H^S(n+1) = f(W^S(n)\times H^S(n)+b^S(n)), n>1; \\
H^S(n) = f(W^S(n)\times C_S+b^S(n)), n=1.\label{eq:1}
\end{aligned}
\end{equation}

\begin{equation}
\begin{aligned}
H^T(n+1) = f(W^T(n)\times H^T(n)+b^T(n)), n>1; \\
H^T(n) = f(W^T(n)\times C_T+b^T(n)), n=1.\label{eq:2}
\end{aligned}
\end{equation}

Here $W^S$ and $b^S$ are parameters from neural network $\Theta^S$, $W^T$ and $b^T$ are parameters from neural network $\Theta^T$. $H^S(n)$ and $H^T(n)$ mean the $n$th hidden layers in the source domain and in the target domain respectively. The two neural networks are first initialized by above functions.

\subsection{Correlation maximization}
To set up the initial relationship of the two neural networks, we resort to Canonical Correlation Analysis (CCA) which can maximize the correlation between two domains \cite{Canonical2004hardoon}. A multi-layer correlation model based on the above deep networks is further constructed. Both the $C_S$ and the $C_T$ are projected by CCA to a common subspace $\Omega$ on which a uniformed representation is generated. Such projection matrices obtained by CCA are denoted as $V^S(n)$ and $V^T(n)$. To find optimal neural networks in the source domain and in the target domain, we have two general objectives: to minimize the reconstruction error of neural networks of the source domain and the target domain, and to maximize the correlation between the two neural networks. To achieve the second objective, we need further on one hand find the best layer matching, on the other hand maximize the correlation between corresponding layers. To achieve this goal, we can minimize the final objective function

\begin{equation}
L({R_{s,t}}) =  \frac{{{L_s}({\theta ^S}) + {L_T}({\theta ^T})}}{{P({V^S},{V^T})}}, \label{eq:3}
\end{equation}
in this function, the objective function is defined as L, and $L({R_{s,t}})$ is in corresponding with different matching of $R_{s,t}$. We would like to generate the best matching by finding the minimum $L({R_{s,t}})$. In $L({R_{s,t}})$, ${L_s}({\Theta^S})$ and ${L_T}({\Theta^T})$ represent the reconstruction errors of data in the source domain and the target domain, which are defined as follows:
\begin{equation}
\begin{aligned}
&{L_S}({\theta ^S}) = [\frac{1}{n_s}\sum\limits_{i = 1}^{n_s} {(\frac{1}{2}||{h_{{W^S},{b^S}}}(C_i^s) - X_i^s|{|^2})} ]\\
& + \frac{\lambda }{2}\sum\limits_{l = 1}^{{n^S} - 1} {\sum\limits_{j = 1}^{n_l^S} {\sum\limits_{k = 1}^{n_{l + 1}^S} {{{(W_{kj}^{S(l)})}^2}} } } \label{eq:4}
 \end{aligned}
\end{equation}

\begin{equation}
\begin{aligned}
&{L_T}({\theta ^T}) = [\frac{1}{n_t}\sum\limits_{i = 1}^{m_t} {(\frac{1}{2}||{h_{{W^T},{b^T}}}(C_i^t) - C_i^t|{|^2})} ]\\
& + \frac{\lambda }{2}\sum\limits_{l = 1}^{{n^T} - 1} {\sum\limits_{j = 1}^{n_l^T} {\sum\limits_{k = 1}^{n_{l + 1}^T} {{{(W_{kj}^{T(l)})}^2}} } },\label{eq:5}
 \end{aligned}
\end{equation}

The third term $P({V_s},{V_t})$ represents the domain divergence after projection by CCA which we want to maximize. The definition for this term is in eq. (\ref{eq:6})

\begin{equation}
P({V^S},{V^T}) = \sum\limits_{l = 2}^{{n^S} - 1} {\frac{{{V^{S{{(l)}^T}}}\sum\nolimits_{ST} {{V^{T(l)}}} }}{{\sqrt {{V^{S{{(l)}^T}}}\sum\nolimits_{SS} {{V^{S(l)}}} } \sqrt {{V^{T{{(l)}^T}}}\sum\nolimits_{TT} {{V^{T(l)}}} } }}}, \label{eq:6}
\end{equation}
where $\sum\nolimits_{ST}  =  {H^{S(l)}}{H^{T{{(l)}^T}}}$ , $\sum\nolimits_{SS}  =  {H^{S(l)}}{H^{S{{(l)}^T}}}$, $\sum\nolimits_{TT}  =  {H^{T(l)}}{H^{T{{(l)}^T}}}$,
By minimizing eq. (\ref{eq:3}), we can collectively train the two neural networks $\theta^T = \{W^T, b^T\}$ and $\theta^S = \{W^S, b^S\}$.

\subsection{Layer matching}
After constructing the multiple layers of the networks by eq. (\ref{eq:3}), we need to further find the best matching for layers after construction of neural networks. As different layer matching would generate different function loss value $L$ in eq. (\ref{eq:3}), we further define the objective function for layer matching as
\begin{equation}
{R_{s,t}} = \arg \min L. \label{eq:7}
\end{equation}

As the solution for eq. (\ref{eq:3}) should be NP-hard, we would like to solve the problem exhaustively. It is also found for regular data, stack auto encoder layers deeper than 5 cannot generate better results, we suppose the layer number on both domains to be lower than 5 here.

\section{Model Training}
Here we would like to first optimize the eq. \ref{eq:3}. As the equation is not joint convex with all the parameters $\theta^S$, $\theta^T$, $V_s$, and $V_t$, and the two parameters $\theta^S$ and $\theta^T$ are not related with $V^S$ and $v^T$, we would like to introduce two-step iteration optimization.

\subsection{Step.1: Updating ${V^S, V^T}$ with fixed ${\Theta^S, \Theta^T}$}
In Eq. (\ref{eq:3}), the optimization of ${V^S, V^T}$ is just related to the dominator term. The optimization of each layer ${V^S(l_1), V^T(l_2)}$(suppose the layer $1_1$ on source domain is in corresponding with layer $l_2$ on target domain) can be formulated as

\begin{equation}
\mathop {\max }\limits_{{V^{S(l_1)}},{V^{T(l_2)}}}  \frac{{{V^{S{{(l_1)}^T}}}\sum\nolimits_{ST} {{V^{T(l_2)}}} }}{{\sqrt {{V^{S{{(l_1)}^T}}}\sum\nolimits_{SS} {{V^{S(l_1)}}} } \sqrt {{V^{T{{(l_2)}^T}}}\sum\nolimits_{TT} {{V^{T(l_2)}}} } }}\label{eq:8}
\end{equation}

As ${{V^{S{{(l_1)}^T}}}\sum\nolimits_{SS} {{V^{S(l_1)}}} }=1$ and ${{V^{T{{(l_2)}^T}}}\sum\nolimits_{TT} {{V^{T(l_2)}}} }=1$ \cite{Canonical2004hardoon}, we can rewrite eq. (\ref{eq:8}) as
\begin{equation}
\begin{array}{l}
\max {V^{S{{(l_1)}^T}}}\sum\nolimits_{ST} {{V^{T(l_2)}}} ,\\
s.t. {V^{S{{(l_1)}^T}}}\sum\nolimits_{SS} {{V^{S(l_1)}}}  = 1, {V^{T{{(l_2)}^T}}}\sum\nolimits_{TT} {{V^{T(l_2)}}}  = 1 \label{eq:9}
\end{array}
\end{equation}
This is a typical constrained problem which can be formulated as a series of unconstrained minimization problems, and be easily solved by Lagrangian multiplier.

\subsection{Step.2: Updating ${\Theta^S, \Theta^T}$ with fixed ${V^S, V^T}$}

As $\Theta^S$ and $\Theta^T$ are mutual independent and with the same form, we here just demonstrate the solution of $\Theta^S$ on the source domain (the solution of $\Theta^T$ can be derived similarly). Actually the objective division operation is with the same function with subtraction operation and we reformulate the objective function as
\begin{equation}
\mathop {\min }\limits_{{\theta ^S}} \phi ({\theta ^S}) = {L_S}({\theta^S}) - \Gamma ({V^S},{V^T})\label{eq:20}
\end{equation}
Here we apply the gradient descent method to adjust the parameter as
\begin{equation}
\begin{aligned}
&{W^{S(l_1)}} = {W^{S(l_1)}} - {\mu ^S}\frac{{\partial \phi }}{{\partial {W^{S(l_1)}}}} \\
&= \frac{{\partial {L_S}({\theta^S})}}{{\partial {W^{S(l_1)}}}} - \frac{{\partial {\Gamma}({V^S},{V^T})}}{{\partial {W^{S(l_1)}}}}\\
&= \frac{{({\alpha ^{S(l_1 + 1)}} - {\beta ^{S(l_1 + 1)}} + {\omega _l}{\gamma ^{S(l_1 + 1)}}) \times {H^{S(l_1)}}}}{{{n_c} + {\lambda ^S}{W^{S(l_1)}}}}\label{eq:21}
\end{aligned}
\end{equation}

\begin{equation}
\begin{aligned}
&{b^{S(l_1)}} = {b^{S(l_1)}} - {\mu ^S}\frac{{\partial \phi }}{{\partial {b^{S(l_1)}}}} \\
&= \frac{{\partial {L_S}({\theta^S})}}{{\partial {b^{S(l_1)}}}} - \frac{{\partial \Gamma ({V^S},{V^T})}}{{\partial {b^{S(l_1)}}}} = \\
&\frac{{({\alpha ^{S(l_1 + 1)}} - {\beta ^{S(l_1 + 1)}} + {\omega _l}{\gamma ^{S(l_1 + 1)}})}}{{{n_c}}},\label{eq:22}
\end{aligned}
\end{equation}
in which
\begin{equation}
{\alpha ^{S(l_1)}} = \left\{ {\begin{array}{*{20}{c}}
{ - ({D_S^l} - {H^{S(l_1)}}) \cdot {H^{S(l_1)}} \cdot (1 - {H^{S(l_1)}}),l = {n^S}}\\
{{W^{S{{(l_1)}^T}}}{\alpha ^{S(l_1 + 1)}} \cdot {H^{S(l_1)}} \cdot (1 - {H^{S(l_1)}})},\\
l = 2,...,{n^S} - 1
\end{array}} \right.
\end{equation}

\begin{equation}
{\beta ^{S(l_1)}} = \left\{ {\begin{array}{*{20}{c}}
{0,\qquad l = {n^S}}\\
{{H^{T(l_2)}}{V^{T(l_2)}}{V^{S{{(l_1)}^T}}} \cdot {H^{S(l_1)}} \cdot (1 - {H^{S(l_1)}})},\\
l = 2,...,{n^S} - 1
\end{array}} \right.
\end{equation}

\begin{equation}
{\gamma ^{S(l_1)}} = \left\{ {\begin{array}{*{20}{c}}
{0,l = {n^S}}\\
{{H^{S(l_1)}}{V^{S(l_1)}}{V^{S{{(l_1)}^T}}} \cdot {H^{S(l_1)}} \cdot (1 - {H^{S(l_1)}})},\\
l = 2,...,{n^S} - 1
\end{array}} \right..
\end{equation}

The operator $\cdot$ here stands for the dot product. The same optimization process works for $\Theta^T$ on the target domain.

\floatname{algorithm}{algorithm}
\renewcommand{\algorithmicrequire}{\textbf{Input:}}
\renewcommand{\algorithmicensure}{\textbf{Output:}}

\begin{CJK*}{UTF8}{gkai}
    \begin{algorithm}
        \caption{Deep Mapping Model Training}\label{alg:1}
        \begin{algorithmic}[1]
            \Require $D^C = \{C_i^s, C_i^t\}_i^{n_c}$,
            \Require $\lambda^S = 1, \lambda^T = 1$, $\mu^S = 0.5, \mu^T = 0.5$
            \Ensure $\Theta(W^S, b^S), \Theta(W^T, b^T), V^S, V^T$
            \Function{NetworkSetUp}{}
            \State Initialize $\Theta(W^S, b^S), \Theta(W^T, b^T)\gets Random Num$
            \Repeat
            \For {{$l=1,2,...,n^S$}}
            \State ${V^S} \gets \arg \min L({\omega _l},{V^{S(l)}})$
            \EndFor
            \For {{$l=1,2,...,n^T$}}
            \State ${V^T} \gets \arg \min L({\omega _l},{V^{T(l)}})$
            \EndFor
            \State ${\theta ^S} = \arg \min \phi ({\theta ^S}), {\theta ^T} = \arg \min \phi ({\theta ^T})$
            \Until {Convergence}
            \EndFunction
            \Function{LayerMatching}{}
            \State Initialize $R_{s,t} \gets Random Matching$
            \State Initialize $m, n \gets 0$
            \If {$m<5$}
            \If {$n<A_5^5/5$}
            \State $R^m_{s,t} = \max {\rm{\{ r}}_n^m{\rm{,r}}_{n + 1}^m{\rm{\} }}$
            \State n = n+1
            \EndIf
            \State $R_{s,t} = \max {\rm{\{ R}}_{s,t}^m{\rm{,r}}_{s,t}^{m + 1}{\rm{\} }}$
            \State m = m+1
            \EndIf
            \EndFunction
        \end{algorithmic}
    \end{algorithm}
\end{CJK*}

After these two optimizations for each layer, the two whole networks (the source domain network and the target domain network) are further fine-tuned by the back-propagation process. The forward and backward propagations will iterate until convergence.

\subsection{Optimization of $R_{s,t}$}
We finally get the minimized $R_{s,t}$ by the above procedures. Take the above layer matching as an example, in  $R_{s,t}$, layer $l_1$ in source domain is in corresponding with
$l_2$ in target domain. As we define the both network for no more than 5 layers(including the original data layers, which will not be used to transfer), the theoretically maximum number of combination should be no more than $A_5^5$(as some layers can be vacancy with no matching). However, in our experiments, we heuristically find the number of matching layers should be in direct proportion to the resolution of images. This observation would save a lot of training time.

The training process is finally summarized in Alg. (\ref{alg:1}), where $r^m_{i,j}$ is generally written as $r^m_n$.

\subsection{Classification on common semantic subspace}
The final classification is performed on the common subspace $\Omega$. The target domain data $D_T$ and the labeled $D_S^l$ are both projected to the common subspace $\Omega$ by the correlation coefficients $V^S(n_S)$ and $V^T(n_T)$. The standard SVM algorithm is applied on $\Omega$. The classifier $\Psi$ is trained by $D_S^l$. This trained classifier $\Psi$ is applied to $D^T$ as $\Psi({D^T}*{V^T})$.

\section{Experiments}

We carry out our DT-LET framework on two cross-domain recognition tasks, handwritten digit recognition, and text-to-image classification.

\subsection{Experimental dataset descriptions}
\textbf{\emph{Handwritten digit recognition:}}For this task, we mainly conduct the experiment on Multi Features Dataset collected from UCI machine learning repository. This dataset consists of features of handwritten numerals (0-9, in total 10 classes) extracted from a collection of Dutch utility maps. 6 features exist for each numeral and we choose the most popular features 216-D profile correlations and 240-D pixel averages in 2*3 windows to complete the transfer learning based recognition task.

\textbf{\emph{Text-to-image classification:}}For this task, we make use of NUS-WIDE dataset. In our experiment, the images in this dataset are represented with 500-D visual features and annotated with 1000-D text tags from Flickr. 10 categories of instances are included in this classification task, which are birds,  building, cars, cat, dog, fish, flowers, horses, mountain, and plane.

\subsection{Comparative methods and evaluation}
As the proposed ET-LET framework mainly have four components, deep learning, CCA, layer matching, and SVM classifier, we first select 3 baseline methods, CCA-SVM \cite{Canonical2004hardoon}, Kernelized-CCA-SVM(KCCA-SVM)\cite{KCCA}, and Deep-CCA-SVM(DCCA-SVM)\cite{Chen2012Marginalized} as baseline comparison methods. We also conduct experiment just without layer matching(the number of layers are the same on the source and the target domains) while all the other parameters are the same with the proposed ET-LET, and we name this framework NoneDT-LET. The final comparison method is the most representative duft-tDTNs method\cite{duft}, which should be up to now heterogenous transfer learning method with best performance.

For the deep network based method, the DCCA-SVM, duft-tDTNs, NoneDT-LET are all with 4 layers for the source domain and the target domain data, as we find more or less layer would generate worse performance.

At last, for the evaluation metric, we select the classification accuracies on the target domain data over the 2 pairs of datasets.
\subsection{Task 1: Handwritten digit recognition}
In the first experiment, we conduct our study for handwritten digit recognition. The source domain data are the 240-D pixel averages in 2*3 windows feature, while the target domain data are the 216-D profile correlations feature. As there are 10 classes in total, we complete 45 ($C_{10}^2$) binary classification tasks, for each category, the accuracy is the average accuracy of 9 binary classification tasks. We use 60\% data as co-occurrence data to complete the transfer learning process and find the common subspace, 20\% labeled samples on source domain as the training samples, and the rest samples on target domain as the testing samples to complete the classification process. The experiments are repeated for 100 times with 100 sets of randomly chosen training and testing data to avoid data bias \cite{Beyond2012Tati}. The final accuracy is the average accuracy of the 100 repeated experiments. This data setting applies for all four methods under comparison.

\begin{figure*}[!ht]
\centering
\includegraphics[width=0.7\linewidth]{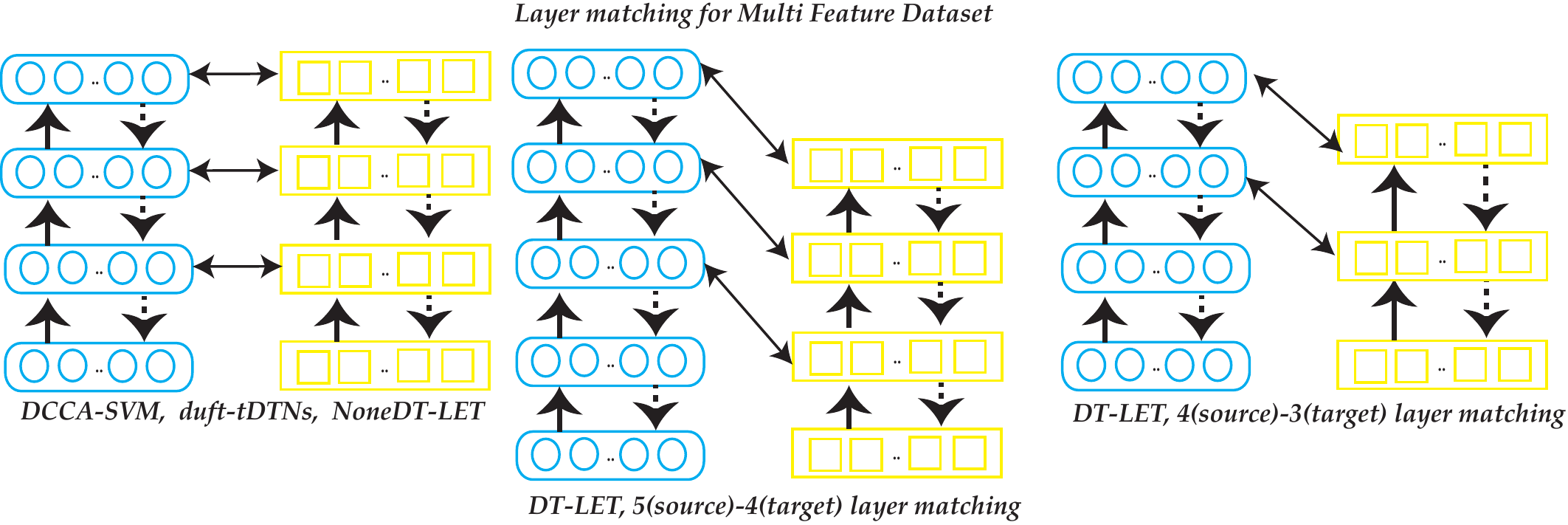}
\caption{The comparison of different layer matching setting for different frameworks on Multi Feature Dataset.}
\label{fig:2}
\end{figure*}

For the deep network, the numbers of neurons of 4 layer networks are 240-170-100-30 for source domain data and 216-154-92-30 for target domain data, this setting works for the all comparison methods. For the proposed DT-LET, we find the best two layer matching with lowest loss after 20 iterations are $r_{4,3}^2$ and $r_{5,4}^3$. The numbers of neurons for $r_{4,3}^2$ are 240-170-100-30 for source domain data and 216-123-30 for target domain data. The average objective function loss of the all 45 binary classification tasks for these two are 0.856 and 0.832 respectively. The numbers of neurons for $r_{5,4}^3$ are 240-185-130-75-30 for source domain data and 216-154-92-30 for target domain data. The one-against-one SVM classification is applied for final classification. The average classification accuracies of 10 categories are shown in Tab. \ref{tab:1}. The matching correlation is detailed in Fig. \ref{fig:2}.

\begin{table*}[t]\footnotesize
\begin{center}
\caption{Classification Accuracy Results on Multi Feature Dataset. (The best performance is emphasized by boldface.)} \label{tab:1}
\begin{tabular}{|c|c|c|c|c|c|c|c|}
  \hline

  numeral & CCA-SVM & KCCA-SVM & DCCA-SVM & duft-tDTNs & NoneDT-LET & DT-LET($r_{5,4}^3$) & DT-LET($r_{4,3}^2$)
  \\
  \hline
  0 & 0.750 & 0.804 & 0.961 & 0.972 & 0.983 & \textbf{0.989} & 0.984\\
  1 & 0.740 & 0.767 & 0.943 & 0.956 & 0.964 & 0.976 & \textbf{0.982}\\
  2 & 0.780 & 0.812 & 0.955 & 0.972 & 0.979 & 0.980 & \textbf{0.989}\\
  3 & 0.748 & 0.790 & 0.945 & 0.956 & 0.966 & \textbf{0.976} & 0.975\\
  4 & 0.752 & 0.799 & 0.956 & 0.969 & 0.980 & \textbf{0.987 }& 0.983\\
  5 & 0.728 & 0.762 & 0.938 & 0.949 & 0.958 & 0.971 & \textbf{0.977}\\
  6 & 0.755 & 0.770 & 0.958 & 0.966 & 0.978 & \textbf{0.988} & 0.986\\
  7 & 0.775 & 0.797 & 0.962 & 0.968 & 0.978 & 0.975 & \textbf{0.985}\\
  8 & 0.764 & 0.793 & 0.948 & 0.954 & 0.965 & 0.968 & \textbf{0.975}\\
  9 & 0.754 & 0.781 & 0.944 & 0.958 & 0.970 & \textbf{0.976} & 0.961\\

  \hline

\end{tabular}
\end{center}
\end{table*}

As can be found in Tab. \ref{tab:1}, the best performances have been highlighted, which all exist in DT-LET framework. However, the best performances for different categories do not exist in the framework with same layer matching. Overall, $r_{5,4}^3$ and $r_{4,3}^2$ should be the best two layer matchings compared with other settings. Based on these results, we heuristically get the conclusion that the best layer matching ratio(5/4, 4/3) is generally in direct proportion to the dimension ratio of original data(240/216). However, more matched layers do not guarantee better performance as the classification results for number ``1", ``2", ``5", ``7", ``8" of DT-LET ($r_{4,3}^2$) with 2 layer matchings perform better than DT-LET($r_{5,4}^3$) with 3 layer matchings.

\subsection{Task 2: Text-to-image classification}
In the second experiment, we conduct our study for Text-to-image classification. The source domain data are the 1000-D text feature, while the target domain data are the 500-D image feature. As there are 10 classes in total, we complete 45 ($C_{10}^2$) binary classification tasks. We still use 60\% data as co-occurrence data\cite{Learning2016Yang}, 20\% labeled samples on source domain as the training samples, and the rest samples on target domain as the testing samples. The same data setting as Task 1 applies for all four methods under comparison.

\begin{figure}[!ht]
\centering
\includegraphics[width=0.8\linewidth]{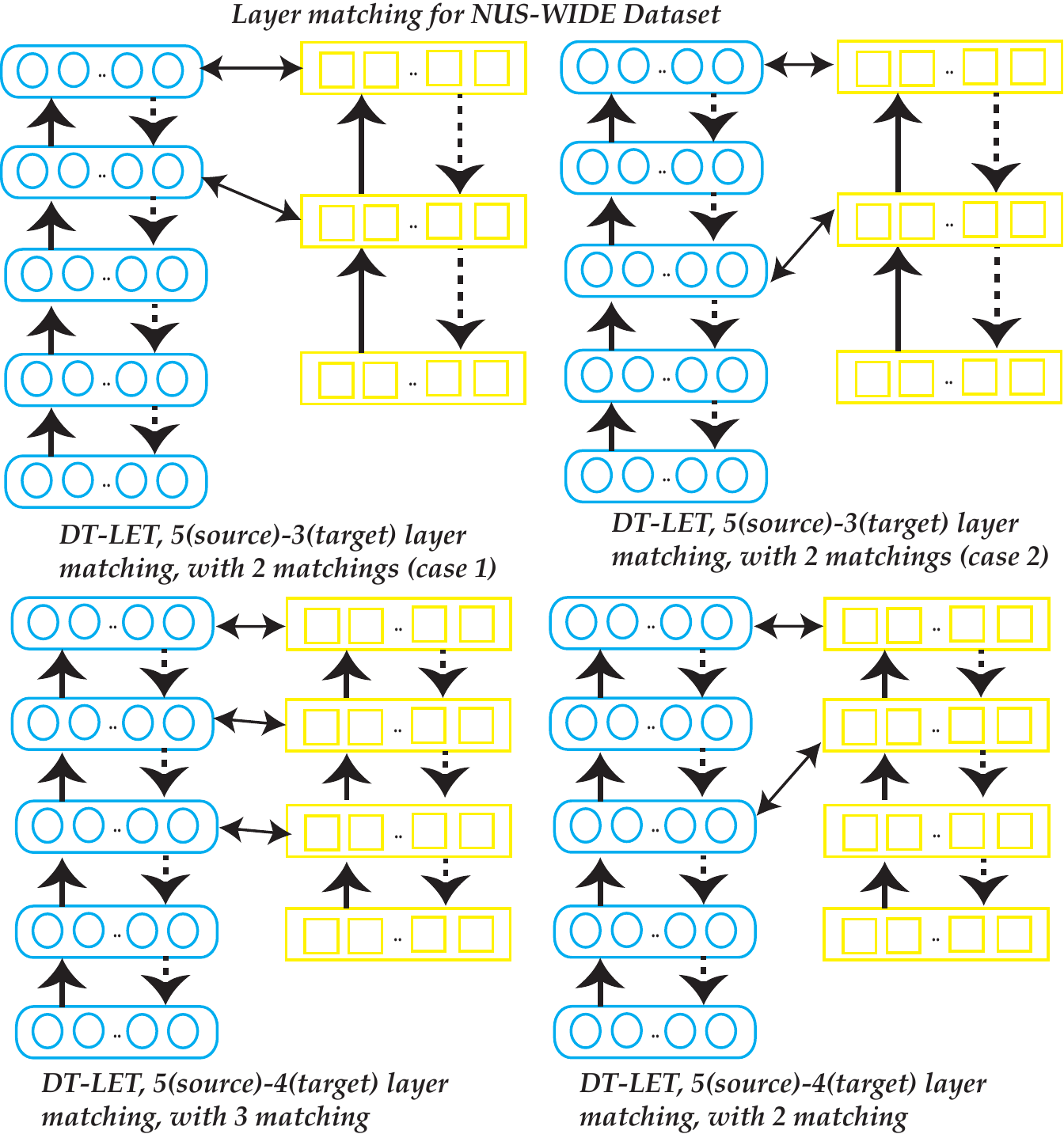}
\caption{The comparison of different layer matching setting for different frameworks on NUS-WIDE Dataset.}
\label{fig:3}
\end{figure}

For the deep network, the numbers of neurons of 4 layer networks are 1000-750-500-200 for source domain data and 500-400-300-200 for target domain data, this setting works for the all comparison methods. For the proposed DT-LET, we find the best two layer matchings with lowest loss after 20 iterations are $r_{5,3}^2$, $r_{5,4}^3$ and $r_{5,4}^2$(non-full rank). The average objective function loss of 45 binary classification tasks for these two layer matchings are 3.231, 3.443 and 3.368. The numbers of neurons for $r_{5,3}^2$ are 1000-800-600--400-200 for source domain data and 500-350-200 for target domain data. The numbers of neurons for both $r_{5,4}^3$ and $r_{5,4}^2$ are 1000-750-500-200 for source domain data and 500-400-300-200 for target domain data. As matching principle would also influence the performance of transfer learning, we present two $r_{5,3}^2$ with different matching principles as shown in Fig. \ref{fig:3}(the  average objective function loss for the two different matching principles are 3.231 and 3.455), in which all the detailed layer matching principles are described.
\begin{table*}[t]\small
\begin{center}
\caption{Classification Accuracy Results on NUS-WIDE Dataset. (The best performance is emphasized by boldface.)} \label{tab:2}
\begin{tabular}{|c|c|c|c|c|c|c|c|c|c|}
  \hline
\multirow{2}*{categories}& \multirow{2}*{CCA-SVM}& \multirow{2}*{KCCA-SVM}& \multirow{2}*{ DCCA-SVM}& \multirow{2}*{duft-tDTNs}& \multirow{2}*{NoneDT-LET}& \multicolumn{4}{c|}{DT-LET}\\
\cline{7-10}
 & & & & & & $r_{5,3}^2$(1) & $r_{5,3}^2$(2) & $r_{5,4}^2$& $r_{5,4}^3$\\
  \hline
  birds & 0.690 & 0.723 & 0.784 & 0.770 & 0.796 & 0.825 & 0.830 &\textbf{ 0.848 } & 0.825\\
  building & 0.706 & 0.741 & 0.810 & 0.783 & 0.816 & 0.881 & 0.838 & 0.881 & \textbf{0.891}\\
  cars & 0.702 & 0.731 & 0.803 & 0.773 & 0.812 & 0.832 & 0.827 & \textbf{0.867} & 0.853\\
  cat & 0.692 & 0.731 & 0.797 & 0.766 & 0.806 & 0.868 & \textbf{0.873} & 0.859 & \textbf{0.873}\\
  dog & 0.687 & 0.726 & 0.798 & 0.765 & 0.805 &0.847 & 0.847 & \textbf{0.863} & 0.823\\
  fish & 0.674 & 0.713 & 0.773 & 0.752 & 0.781 & 0.848 & 0.834 & \textbf{0.852} & 0.839\\
  flowers & 0.698 & 0.733 & 0.799 & 0.783 & 0.805 & 0.863 & 0.844 & 0.844 & \textbf{0.875}\\
  horses & 0.700 & 0.736 & 0.802 & 0.775 & 0.808 & \textbf{0.841 }& 0.812 & \textbf{0.841} & 0.831\\
  mountain & 0.717 & 0.748 & 0.816 & 0.786 & 0.827 & 0.825 & 0.813 & 0.821 & \textbf{0.831}\\
  plane & 0.716 & 0.747 & 0.824 & 0.787 & 0.828 & 0.810  & \textbf{0.832} & \textbf{0.832} & 0.825\\
  average & 0.698 & 0.733 & 0.801 & 0.774 & 0.808 & 0.844 & 0.833 & \textbf{0.851} & 0.847 \\
  \hline

\end{tabular}
\end{center}
\end{table*}
For this task, as the overall accuracies are generally lower than task 1, we would like to compare more different settings for this cross-layers matching task. We first verify the effectiveness of DT-LET framework. Compared with the comparison methods, the accuracy of DT-lET framework is generally with around 85\% accuracy while the comparison methods are generally with no more than 80\%. This observation generates the conclusion that finding the appropriate layer matching is essential. The second comparison is between the full rank and non-full rank framework. As can be found in the table, actually $r_{5,4}^2$ is with the highest overall accuracy, although the other non-full rank DT-LETs do not perform quite well. This observation gives us a hint that full rank transfer is not always best as negative transfer would degrade the performance. However, the full rank transfer is generally good, although not optimal. The third comparison is between the same transfers with different matching principles. We present two $r_{5,3}^2$ with different matching principles, and we find the performances vary. The case 1 performs better than case 2. This result tell us continuous transfer might be better than discrete transfer: as for case 1, the transfer is in the last two layers of both domains, and in case 2, the transfer is conducted in layer 3 and layer 5 of the source domain data.

By comparing specific objects, we can find the objects with large semantic difference with other categories are with higher accuracy. For the objects which are hard to classify and with low accuracy, like ``birds" and ``plane", the accuracies are always low even the DT-LET is introduced. This observation proves the conclusion that DT-LET can only be used to improve the transfer process, which helps with the following classification process; while the classification accuracy is still based on the semantic difference of data of different categories.

We also have to point out the relationship between the average objective function loss and the classification accuracy is not strictly positive correlated. Overall $r_{5,4}^2$ is with the highest classification accuracy while its average objective function loss is not lowest. Based on this observation, we have to point out, the lowest average objective function loss can only generate the best transfer leaning result with optimal common subspace. On such common subspace, the data projected from target domain are classified. These classification results are also influenced by the classifier as well as training samples projected randomly from the source domain. Therefore, we conclude as follows. We can just guarantee a good classification performance after getting the optimal transfer learning result, while the classification accuracy is also influenced by the classification settings.

\subsection{Parameter sensitivity}
In this section, we study the effect of different parameters in our networks. We have to point out the even the layer matching is random, the last layer of the two neural networks from the source domain and the target domain must be correlated to construct the common subspace.  Actually, the number of neurons at last layer would also affect the final classification result. For the last layer, we take experiments on Multi Feature Dataset as an example. The result is shown in Tab. \ref{tab:7}.

From this figure, it can be noted when the number of neuron is 30, the performance is the best. Therefore in our former experiments, 30 neurons are used. The conclusion can also be drawn that more neurons are not always better. Based on this observation, The number of layers in Task 1 is set as 30, and in Task 2 as 200.

\begin{table}[t]\small
\begin{center}
\caption{Effects of the number of neurons at the last layer} \label{tab:7}
\begin{tabular}{|c|c|c|c|c|c|}

  \hline
 layer matching & 10 & 20 & 30 & 40 & 50\\

  \hline
$r_{5,4}^3$ & 0.9082& 0.9543& 0.9786& 0.9771& 0.9653  \\
  \hline
$r_{4,3}^2$ & 0.8853& 0.9677& 0.9797& 0.9713& 0.9522  \\
  \hline
\end{tabular}
\end{center}
\end{table}

\section{Conclusion}

In this paper, we propose a novel framework, referred to as Deep Transfer Learning by Exploring where to Transfer(DT-LET), for hand writing digit recognition and text-to-image classification. In the proposed model, we find the best matching with lowest loss value. After the transfer, the final correlated common subspace on which classifier is applied. Experimental results support the effectiveness of the proposed framework.

As the current framework is only suitable for binary classification, extending it to multi-class classification is our future work. We would also propose more robust model to solve this ``where to transfer" problem in the future.
\bibliographystyle{aaai}
\bibliography{ref}
\end{document}